\documentclass[letterpaper]{article} 
\usepackage[preprint]{aaai2027}  
\usepackage[hyphens]{url}  
\usepackage{graphicx} 
\usepackage{natbib}  
\usepackage{caption} 
\usepackage{amsmath}
\usepackage{amsthm}
\usepackage{amssymb}
\usepackage{algorithm}
\usepackage{algorithmic}
\usepackage{subcaption}

\usepackage{newfloat}
\usepackage{listings}
\DeclareCaptionStyle{ruled}{labelfont=normalfont,labelsep=colon,strut=off} 
\floatstyle{ruled}
\newfloat{listing}{tb}{lst}{}
\floatname{listing}{Listing}

\usepackage{booktabs}

\title{GSLAD: Prototype-Regularized Graph Structure Learning for \\Multivariate Time Series Anomaly Detection}
\author{
    Zepeng Zhang, Fuad Khuri, Keivan Faghih Niresi, Olga Fink\corresponding
}
\affiliations{

    Intelligent Maintenance and Operations Systems (IMOS) Lab

    \'{E}cole Polytechnique F\'ed\'erale de Lausanne (EPFL), Lausanne, Switzerland\\
    \{zepeng.zhang, fuad.khuri, keivan.faghihniresi, olga.fink\}@epfl.ch
}

\begin{document}

\maketitle

\begin{abstract}
Unsupervised multivariate time series anomaly detection methods typically identify anomalies through forecasting, reconstruction, or representation discrepancies. However, industrial faults may first alter inter-variable structural patterns while individual trajectories remain close to normal, resulting in weak anomaly signals. In this paper, we propose GSLAD, a prototype-regularized graph structure learning framework that uses structural deviations for anomaly scoring. GSLAD adopts a two-phase training strategy. First, a condition-aware graph learner and a graph-based forecaster are optimized with predictive supervision. The inferred normal graphs are then clustered into multiple structural prototypes representing different normal operating regimes, with edge-wise variability characterizing structural uncertainty. Deviations from these prototypes regularize the graph learner in the second phase, encouraging stable and regime-specific structural patterns. During inference, uncertainty-normalized structural deviation is combined with predictive deviation for anomaly scoring. Experiments on four industrial benchmarks demonstrate strong overall performance of GSLAD and confirm the effectiveness of structural deviation for anomaly detection and diagnosis.
\end{abstract}

\section{Introduction}
Many industrial systems are monitored with sensor networks that generate vast amounts of multivariate time series (MTS) data.
Detecting anomalies in MTS data is essential for preventing failures, reducing downtime, and maintaining the reliable operation of complex industrial systems \cite{wu2021graph,deng2021graph,fink2026physics}.
In practice, however, fault labels are often scarce, incomplete, and expensive to obtain, motivating increasing interest in unsupervised MTS anomaly detection \cite{zhang2019deep,audibert2020usad,belay2023unsupervised}.

Unsupervised MTS anomaly detection models are typically trained to forecast \cite{deng2021graph} or reconstruct \cite{zhao2024dyedgegat} normal observations.
Then, during inference, anomalies are identified by measuring discrepancies between observed signals and model outputs \cite{jin2024survey,chen2024graph,ho2025graph}.
Despite their architectural differences, most existing methods rely on a similar assumption: anomalous samples are expected to produce larger predictive, reconstructive, or representation discrepancies than normal samples \cite{zhao2020multivariate,cho2025structured}. 
This principle is effective when faults cause pronounced deviations in individual sensor trajectories.
However, many industrial faults do not immediately induce large marginal deviations.
Instead, they may primarily alter the structural patterns while individual sensor trajectories remain close to their normal ranges.
As a result, a model may continue to forecast or reconstruct individual trajectories accurately, producing weak anomaly signal.

To capture structural patterns of the system, several MTS anomaly detection methods propose to incorporate graph structure learning into graph neural network (GNN)-based forecasting or reconstruction models \cite{deng2021graph,zheng2023correlation,zhao2024dyedgegat}. In these approaches, however, the learned graph primarily serves as an intermediate computational structure for more effective message passing, while anomaly scores are still derived solely from prediction or reconstruction discrepancies. More recently, several methods have explicitly exploited structural information for anomaly detection, for example by constructing relational graphs from model gradients \cite{liu2025gcad} or intermediate representations \cite{cho2025structured}. Nonetheless, the graph structures in these works used for anomaly scoring are derived post hoc from trained predictors or latent representations, rather than being jointly inferred as explicit model outputs.

Turning structural deviation into a reliable anomaly signal is nontrivial.
Structural patterns may vary across different operating regimes, and complex systems often exhibit multiple distinct normal operating regimes.
Moreover, normal structural variability is heterogeneous across edges.
Deviations on stable edges should provide stronger anomaly evidence than deviations on edges that fluctuate naturally during normal operation.
To address these challenges, we introduce \textit{GSLAD}, a prototype-regularized graph structure learning framework for unsupervised MTS anomaly detection.
GSLAD jointly infers condition-dependent graph structures and forecasts future observations with a GNN.
Its two-phase training procedure first learns predictive normal graphs and then summarizes them into multiple uncertainty-aware structural prototypes. These prototypes are expected to capture distinct normal operating regimes, while their edge-wise variability quantifies the reliability of individual structural relations. Deviation from these prototypes subsequently regularizes the graph learner, encouraging normal graphs to remain compact around regime-specific prototypes. During inference, uncertainty-normalized structural deviation from the nearest normal prototype is combined with predictive deviation for anomaly scoring.

The main contributions are summarized as follows:
\begin{itemize}
    \item We propose GSLAD and develop a two-phase training strategy. The first phase learns condition-aware graphs under predictive supervision, while the second phase regularizes graph learning using structural prototypes, thereby capturing stable and regime-specific structural patterns. 

    \item Beyond conventional residual-based anomaly scoring, we perform anomaly detection in a jointly learned structural space, where uncertainty-aware structural deviation is combined with predictive deviation for anomaly scoring.

    \item Experiments on four industrial benchmarks demonstrate the strong overall anomaly detection performance of GSLAD. The structural analysis and fault characterization demonstrate the potential for further fault diagnosis.
\end{itemize}

\section{Related Work}
\subsection{Multivariate Time-Series Anomaly Detection}

Unsupervised MTS anomaly detection methods typically learn normal system behavior and identify anomalies according to deviations from the learned normality.
Most existing methods derive anomaly scores from prediction or reconstruction residuals.
For example, MTAD-GAT \cite{zhao2020multivariate} jointly optimizes forecasting and reconstruction objectives on normal data and computes anomaly scores from the discrepancies between observations and model outputs.
Graph-based methods further model inter-variable dependencies to improve forecasting or reconstruction.
GDN \cite{deng2021graph} learns sensor relationships for forecasting-based anomaly detection, whereas DyEdgeGAT \cite{zhao2024dyedgegat} infers input-dependent graph structures for reconstruction-based anomaly detection.
Beyond value-space residuals, other methods detect anomalies through association discrepancies \cite{xu2022anomaly}, representation discrepancies \cite{yang2023dcdetector}, or frequency-domain deviations \cite{wu2025catch}.
Despite their architectural differences, these methods primarily quantify abnormality through deviations in observed values or latent representations.
They may therefore provide weak anomaly evidence when faults alter structural patterns without immediately inducing pronounced deviations in individual trajectories.

More closely related to GSLAD, several recent methods exploit structural changes for anomaly detection.
Specifically,
GRELEN \cite{zhang2022grelen} characterizes anomalies through changes in the in-degree and out-degree distributions of learned relational graphs,
GCAD \cite{liu2025gcad} detects anomalies from changes in relational graphs derived from the gradients of a trained predictor, while OracleAD \cite{cho2025structured} constructs relational graphs from intermediate representations and measures their structural deviations for anomaly scoring.
Unlike methods that use graphs primarily as message-passing topologies or derive structural evidence from a trained model, GSLAD explicitly measures structural deviations in a jointly learned structural space.

\subsection{Graph Structure Learning}

Graph structure learning aims to infer or refine graph topologies directly from data rather than relying exclusively on predefined structures \cite{zhu2021survey,li2023gslb}.
Existing graph structure learning methods have been developed to address problems where observed graphs are noisy, incomplete, or unavailable \cite{jin2020graph,liu2022towards,zhang2024graph}.
They typically optimize the learned graph structure together with node representations for downstream tasks such as classification, prediction, or representation learning.
Accordingly, the graph mainly functions as an intermediate computational structure that determines how information is propagated by a GNN.
GSLAD differs from conventional graph structure learning methods in that it not only uses the learned graph structure for more effective message passing but also explicitly perform anomaly detection in the learned structural space.

\section{Prototype-Regularized\\ Graph Structure Learning}
\begin{figure*}
    \centering
    \includegraphics[width=1.0\textwidth]{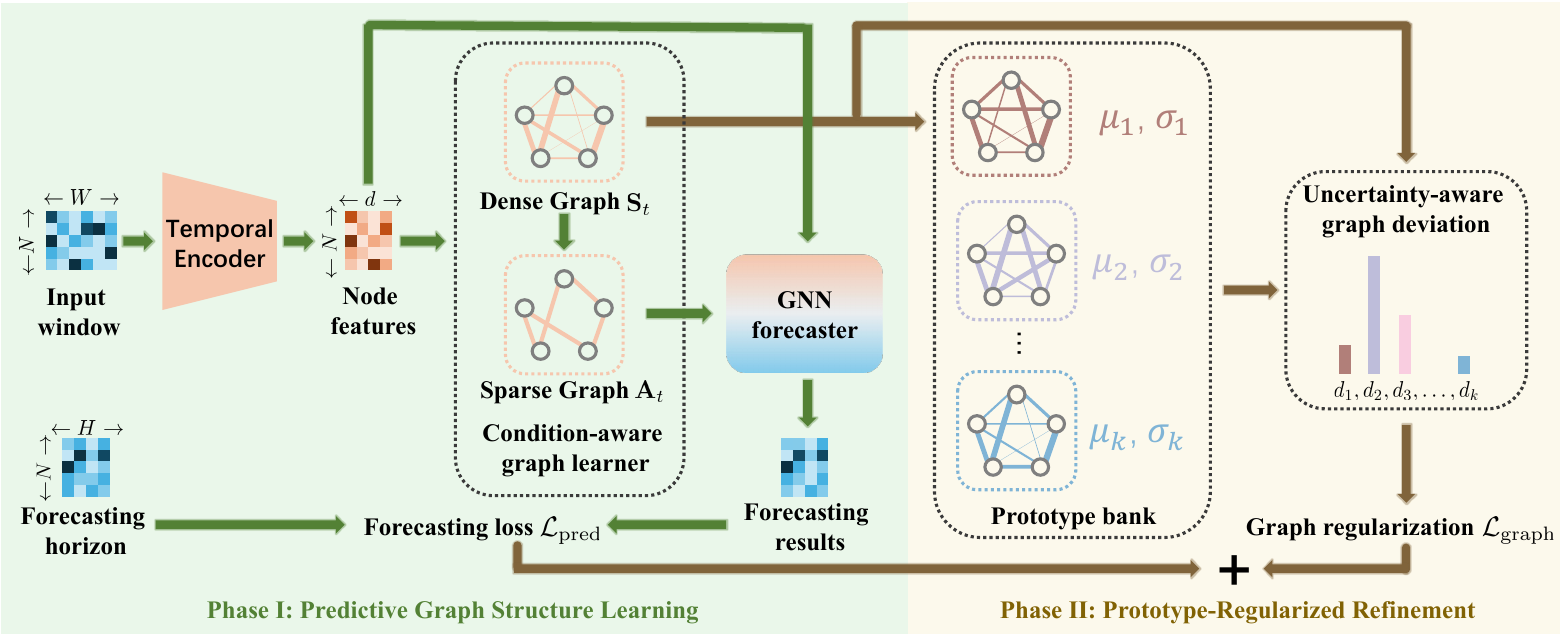}
    \caption{Overview of GSLAD. In Phase I, the temporal encoder, condition-aware graph learner, and GNN forecaster are optimized using predictive supervision. Dense normal graphs are clustered into multiple uncertainty-aware prototypes, while sparsified graphs are used for efficient GNN processing. In Phase II, deviation from the prototype bank regularizes graph learning. During inference, uncertainty-aware structural deviation and predictive deviation are combined for anomaly scoring.}
    \label{fig:framework}
\end{figure*}

In this section, we first introduce the GSLAD architecture, which consists of a temporal encoder, a condition-aware graph learner, and a GNN forecaster.
Then we introduce the two-phase training strategy and the anomaly scoring policy.
A schematic overview of the GSLAD framework is provided in Figure~\ref{fig:framework}.

\subsection{Predictive Dynamic Graph Learning}
We denote by $\mathbf{X}_t$ a window containing past $W$ observations from $N$ variables:
\begin{equation}
    \mathbf{X}_t = [\mathbf{x}_{:,t-W+1}, \dots, \mathbf{x}_{:,t}] 
    \in \mathbb{R}^{N \times W},
\end{equation}
and by $\mathbf{Y}_{t}$ the subsequent $H$-step forecasting target:
\begin{equation}
    \mathbf{Y}_{t} = [\mathbf{x}_{:,t+1}, \dots, \mathbf{x}_{:,t+H}] \in \mathbb{R}^{N \times H}.
\end{equation}
Given $\mathbf{X}_t$, GSLAD predicts $\hat{\mathbf{Y}}_{t}$ and simultaneously infers a dense directed graph $\mathbf{S}_t\in[0,1]^{N\times N}$.
The model consists of a temporal encoder, a dynamic graph learner, and a GNN forecaster.
The details of these three components are introduced in the following.
\subsubsection{Temporal Encoding}
The historical window of each variable is encoded independently using a shared temporal encoder.
Specifically, for input $\mathbf{X}_{t}=[\mathbf{x}_{1,:},\ldots,\mathbf{x}_{N,:}]^\top$, we use an one-dimensional convolutional neural network \cite{kiranyaz20211d} to encode it into latent space as follows:
\begin{equation}
    \mathbf{h}_{i,:}
    =
    \operatorname{Conv1D}(\mathbf{x}_{i,:})
    \in \mathbb{R}^{d},
\end{equation}
yielding node representations
\begin{equation}
    \mathbf{H}_t =
    [\mathbf{h}_{1,:}, \ldots, \mathbf{h}_{N,:}]^\top
    \in \mathbb{R}^{N \times d},
\end{equation}
where $d$ is the representation dimension. Sharing the encoder across nodes provides a consistent representation space while preserving node-specific temporal patterns.
These node representations will then be used as inputs for both the following dynamic graph learner and the GNN forecaster.

\subsubsection{Condition-Aware Graph Inference}
Given the node representations $\mathbf{H}_t$, the graph learner constructs a dynamic graph through an attention-based module.
Considering that the normal operation state of a system normally contains multiple operating conditions, we propose a condition-aware attention mechanism for graph learning. 
Specifically, we modulate node-wise query and key projections using a global state representation computed by mean-pooling all node representations as follows:
\begin{equation}
    \mathbf{c}_t
    =
    \frac{1}{N}
    \sum_{i=1}^{N}
    \mathbf{h}_{t,i}.
\end{equation}
This mechanism allows the same variable pair to receive different affinities under different operating conditions.
The global state vector $\mathbf{c}_t$ is used to modulate the query projection $\mathbf{q}_{t,i}$ and key projection $\mathbf{k}_{t,i}$ of each
node:
\begin{equation}
\begin{aligned}
&\mathbf{q}_{t,i}
    =
    \left(
    \mathbf{W}_{q} \mathbf{h}_{t,i}
    \right)
    \odot
    \sigma
    \left(
    \mathbf{W}^{c}_{q} \mathbf{c}_t
    \right),\\
    &\mathbf{k}_{t,i}
    =
    \left(
    \mathbf{W}_{k} \mathbf{h}_{t,i}
    \right)
    \odot
    \sigma
    \left(
    \mathbf{W}^{c}_{k} \mathbf{c}_t
    \right),    
\end{aligned}  
\end{equation}
where $\odot$ denotes element-wise multiplication, $\mathbf{W}_{q}$, $\mathbf{W}^{c}_{q}$, $\mathbf{W}_{k}$, and $\mathbf{W}^{c}_{k}\in\mathbb{R}^{d\times d}$ are weight matrices shared across nodes. 
The edge weight for each node pair $(i,j)$ is computed as follows:
\begin{equation}
\begin{aligned}
    e_{t,ij}
    &=\frac{
    \mathbf{q}_{t,i}^{\top}
    \mathbf{k}_{t,j}
    }{
    \sqrt{d}
    },\\
    s_{t,ij}
    &=
    \sigma(e_{t,ij})\in (0,1),
\end{aligned}
\end{equation}
where $\sigma$ represents the sigmoid function to map the pairwise score to $(0,1)$.
The resulting dense graph $\mathbf{S}_t$ is retained for the following prototype construction, which will be used to regularize graph learning during training and structural anomaly scoring during testing.

\subsubsection{Graph-Based Forecasting}
With the learned dynamic graph, we use a GNN model to perform MTS forecasting.
To facilitate efficient computation, we first convert the dense adjacency matrix into a sparse weighted graph $\mathbf{A}_t$.
Specifically, we retain the top-$k$ outgoing neighbors of each node and apply a masked softmax over the retained logits, which is computed by
\begin{equation}
a_{t,ij} =
\begin{cases}
\dfrac{\exp(e_{t,ij})}
{\sum_{m \in \operatorname{TopK}(e_{t,i:})} \exp(e_{t,im})},
& j \in \operatorname{TopK}(e_{t,i:}), \\
0,
& \text{otherwise}.
\end{cases}
\end{equation}
We distinguish the sparse graph $\mathbf{A}_t$, used only for message passing, from the dense graph $\mathbf{S}_t$, used for prototype learning and anomaly scoring.
The sparse graph $\mathbf{A}_t$ is used by the GNN forecaster to efficiently propagate information across variables.
With $\mathbf{Z}^{(0)}_t = \mathbf{H}_t$, each layer updates node representations using
\begin{equation}
    \mathbf{Z}^{(\ell+1)}_t
    =
    \operatorname{GNN}^{(\ell)}
    \left(
    \mathbf{Z}^{(\ell)}_t, \mathbf{A}_t
    \right),\quad\ell=1,\ldots,L-1,
\end{equation}
where $\operatorname{GNN}^{(\ell)}(\cdot)$ denotes one GNN layer, including a message passing step followed by nonlinearity.
In our implementation, we use a GAT-v2 layer \cite{brody2022how}, while the framework also supports other GNN layer options. 
After $L$ layers of message passing, a node-wise MLP decodes the final representation to the forecasting horizon as follows:
\begin{equation}
    \hat{\mathbf{y}}_{t,i}
    =
    \mathrm{MLP}
    \left(
    \mathbf{z}^{(L)}_{t,i}
    \right)
    \in \mathbb{R}^{H}.
\end{equation}

\subsection{Prototype-Regularized Two-Phase Training}
GSLAD follows a two-phase training procedure. 
Phase I learns predictive dependency structures using forecasting supervision. 
The dense graphs inferred from normal training windows are then summarized into a prototype bank, which serves as the structural reference for Phase II. 
Phase II training further refines the graph learner using both forecasting supervision and prototype-based regularization.

\paragraph{Phase I: Predictive Graph Learning}
Following the standard setting in graph structure learning literature \cite{zhu2021survey,li2023gslb}, we alternate between updating the graph learner while freezing the forecaster and updating the forecaster while freezing the graph learner.
Both steps are optimized with a unified predictive objective:
\begin{equation}
    \mathcal{L}_{\mathrm{pred}}
    =
    \frac{1}{NH}
    \left\|
    \hat{\mathbf{Y}}_t - \mathbf{Y}_t
    \right\|_F^2 .
\label{eq:prediction_mse_loss}
\end{equation}
This phase allows the temporal encoder, graph learner, and GNN forecaster to learn predictive structural patterns.
Note that only the parameters in the graph learner and GNN forecaster are optimized alternately, while the encoder is optimized in both alternating stages.
\paragraph{Normal Graph Prototype Construction}
After Phase I, we collect the learned dense graphs $\{\mathbf{S}_t\}_{t \in \mathcal{T}}$, where $\mathcal{T}$ denotes the training set windows. 
A straightforward strategy is to use the average graph as the prototype to regularize the graph learning and to compute the structural deviation. 
However, interpolating between distinct operating regimes may produce a prototype graph that corresponds to no actual operating regime, and comparing every sample with such an averaged graph may incorrectly penalize normal structural changes.
Therefore, we assume that normal graph structures concentrate around a small number of operating-regime-dependent prototypes. 
Specifically, we cluster the dense graphs inferred from normal training windows into $K$ prototypes using $K$-means. 
For $k$-th prototype graph, we compute the edge-wise mean and empirical standard deviation within the corresponding cluster:

\begin{equation}
\begin{aligned}
\mu_{k,ij}
&=
\frac{1}{|\mathcal{T}_k|}
\sum_{t\in\mathcal{T}_k}s_{t,ij},
\\
\sigma_{k,ij}
&=
\sqrt{
\frac{1}{|\mathcal{T}_k|}
\sum_{t\in\mathcal{T}_k}
\left(s_{t,ij}-\mu_{k,ij}\right)^2
}.    
\end{aligned}
\end{equation}
where $\mathcal{T}_k$ is the set of windows assigned to the $k$-th prototype. 
The mean graph $\boldsymbol{\mu}_k$ represents the structural pattern of regime $k$, while $\boldsymbol{\sigma}_k$ captures edge-wise variability within that regime.
The resulting set $\{(\boldsymbol{\mu}_k,\boldsymbol{\sigma}_k)\}_{k=1}^K$ forms the normal graph prototype bank used in prototype-regularized refinement and structural deviation computation.
These standard deviations $\sigma_{k,ij}$ will then be used to weight the deviation of edge $(i,j)$.
Specifically, a small $\sigma_{k,ij}$ indicates a stable and clear structural pattern and therefore should receive larger weights.
For edges with large $\sigma_{k,ij}$, we assign lower weights as they fluctuate naturally even during normal operation conditions.
 
\paragraph{Phase II: Prototype-Regularized Refinement}
With the constructed normal graph prototype bank, we perform graph refinement by augmenting the loss in \eqref{eq:prediction_mse_loss} with an additional prototype-regularized refinement term, while the GNN forecaster is trained with the same predictive objective in \eqref{eq:prediction_mse_loss}.
Specifically, for a learned graph $\mathbf{S}_t$, we measure its uncertainty-normalized graph deviation for each prototype graph as follows:
\begin{equation}
    d_k(\mathbf{S}_t)
    =
    \frac{1}{N^2}
    \sum_{i=1}^N \sum_{j=1}^N
    \frac{
    (s_{t,ij} - \mu_{k,ij})^2
    }{
    \sigma_{k,ij}^2 + \sigma_0^2
    },
\end{equation}
where $\sigma_0$ is a small global stabilization term preventing near-zero variances from dominating the objective.
This uncertainty-aware graph deviation is implemented as a diagonal Mahalanobis-style distance that assigns larger penalties to deviations on stable edges and smaller penalties to deviations on edges with naturally large variations.
During training, we use a soft nearest prototype graph assignment as follows:
\begin{equation}
    w_{t,k}
    =
    \frac{
    \exp(-d_k(\mathbf{S}_t)/\tau)
    }{
    \sum_{j=1}^{K}
    \exp(-d_j(\mathbf{S}_t)/\tau)
    },
\end{equation}
where $\tau$ is a temperature parameter. The prototype graph regularization is computed as
\begin{equation}
    \mathcal{L}_{\mathrm{graph}}
    =
    \sum_{k=1}^{K}
    w_{t,k} d_k(\mathbf{S}_t).
\end{equation}
During training, the gradients only back-propagate through the graph distance terms but are stopped through the weights $w_{t,k}$, so the soft weights act only as prototype-selection coefficients rather than being optimized to trivially reduce the regularization objective.
In summary, the training loss during Phase II is defined by
\begin{equation}
    \mathcal{L}
    =
    \mathcal{L}_{\mathrm{pred}}
    +
    \lambda
    \mathcal{L}_{\mathrm{graph}} .
\end{equation}
Note that the prototype bank obtained in Phase I remains fixed throughout Phase II, preventing the reference distribution from drifting together with the graph learner.
After Phase II training, we recompute the prototype graph bank and the corresponding statistics based on the newly obtained inferred graphs, forming the refined prototype graph bank, which is then used for structural deviation computing during inference.
\begin{table*}
\begin{centering}
\par\end{centering}
\centering{}\resizebox{1 \textwidth}{!}{%
\begin{tabular}{ccccccccc}
\toprule 
 & \multicolumn{2}{c}{SWAT}  & \multicolumn{2}{c}{WADI} & \multicolumn{2}{c}{PSM} & \multicolumn{2}{c}{TEP} \tabularnewline
 & AUC-ROC & AUC-PR & AUC-ROC & AUC-PR & AUC-ROC & AUC-PR & AUC-ROC & AUC-PR \tabularnewline
 \midrule 
 MTAD-GAT & 0.7619 & 0.2280 & 0.6105 & 0.2284 & $\underline{\text{0.7885}}$ & 0.5464 & 0.9021 & 0.9136 \tabularnewline
 GDN & 0.7360 & 0.1975 & 0.6059 & 0.2276 & 0.7568 & $\underline{\text{0.5812}}$ & 0.8990 & 0.9130 \tabularnewline
 GRELEN & 0.7708 & 0.2320 &0.5848 & 0.2282& 0.6189 & 0.4003 & 0.8190 & 0.8355 \tabularnewline
 Anomaly Transformer & 0.7605 & 0.2367 & 0.4940 & 0.1590 & 0.7202 & 0.5391 & $\underline{\text{0.9326}}$ & 0.9494 \tabularnewline
 NLinear & 0.8187 & 0.7214 & 0.7086 & 0.4469 & 0.6843 & 0.5482 & 0.7382 & 0.7874 \tabularnewline
 DLinear & 0.5931 & 0.1263 & 0.6993 & 0.4462 & 0.6661 & 0.5237 & 0.8006 & 0.8201 \tabularnewline
 DCDetector & 0.7359 & 0.6264 & 0.6929 & 0.4442 & 0.6513 & 0.4896 & 0.8862 & 0.9131 \tabularnewline
 PatchTST & 0.8149 & 0.7198 & 0.7011 & 0.4458 & 0.6722 & 0.5071 & 0.8248 & 0.8667 \tabularnewline
 DyEdgeGAT & 0.7394 & 0.2406 & 0.6517 & 0.4426 & 0.6976 & 0.5407 & 0.9325 & $\underline{\text{0.9510}}$ \tabularnewline
 iTransformer & $\underline{\text{0.8213}}$ & $\underline{\text{0.7231}}$ & $\underline{\text{0.7129}}$ & 0.4482 & 0.6576 & 0.5075 & 0.7346 & 0.7436 \tabularnewline
 ModernTCN & 0.7555 & 0.6559 & 0.7064 & 0.4512 & 0.7074 & 0.5453 & 0.8903 & 0.9166 \tabularnewline
 CATCH & 0.8110 & 0.7178 & 0.7112 & 0.4467 & 0.6710 & 0.5056 & 0.8575 & 0.8868 \tabularnewline
 GCAD & 0.5597 & 0.1196 & \textbf{0.7197} & $\underline{\text{0.4490}}$ & 0.6611 & 0.4899 & 0.8500 & 0.8820 \tabularnewline
GSLAD & \textbf{0.8607} & \textbf{0.7612} & 0.7018 & \textbf{0.5157} & \textbf{0.8193} & \textbf{0.7038} & \textbf{0.9563}  & \textbf{0.9687}\tabularnewline
\bottomrule
\end{tabular}}
\caption{Multivariate time series anomaly detection results.}
\label{tab:mtsad_results}
\end{table*}
\subsection{Structural and Predictive Anomaly Scoring}
Since the GSLAD model performs graph learning and GNN forecasting simultaneously, it produces two complementary anomaly scores, namely, a predictive deviation and a structural deviation.
The predictive deviation measures the distance between the forecasted time series and the real time series:
\begin{equation}
    r_t
    =
    \left\|
    \hat{\mathbf{Y}}_t - \mathbf{Y}_t
    \right\|_F^2 .
\end{equation}
The uncertainty-aware structural deviation measures the distance between the learned graph and the closest prototype graph:
\begin{equation}
    d_t=d(\mathbf{S}_t)
    =
    \min_{k \in \{1,\ldots,K\}}
    \frac{1}{N^2}
    \sum_{i=1}^N \sum_{j=1}^N
    \frac{
    (\mathbf{s}_{t,ij} - \boldsymbol{\mu}_{k,ij})^2
    }{
    \boldsymbol{\sigma}_{k,ij}^2 + \sigma_0^2
    }.
\end{equation}
We use the minimum over different prototype graphs instead of the average because a test window should be considered as normal if its behavior is close to any learned normal operating condition.
Since structural deviation and predictive deviation may have different scales, we use the sum of $z$-score normalized deviations as the anomaly score.

\section{Experiments}
In this section, we evaluate the effectiveness of the proposed GSLAD approach for MTS anomaly detection. 
First, we introduce the experimental settings. 
Then, we assess the effectiveness of GSLAD on anomaly detection tasks on four industrial datasets and we conduct ablation studies and parameter sensitivity analysis to investigate the contributions of individual components in GSLAD.
Finally, we analyze the structural deviation patterns to perform anomaly diagnosis.

\subsection{Experiment Settings}
\subsubsection{Datasets}
We conduct experiments on four industrial datasets, namely, the Secure Water Treatment (SWaT) dataset \cite{mathur2016swat}, the Water Distribution (WADI) dataset \cite{ahmed2017wadi}, the Pooled Server Metrics (PSM) dataset \cite{abdulaal2021practical}, and the Tennessee Eastman Process (TEP) dataset \cite{reinartz2021extended}.
SWaT and WADI contain sensor and actuator measurements collected from water-treatment and water-distribution testbeds, respectively. PSM consists of server-level monitoring metrics, while TEP simulates a multivariable chemical process with multiple fault types. 

\subsubsection{Baselines}
We compare the proposed GSLAD model with thirteen representative baselines covering forecasting-based, reconstruction-based, representation-based, and graph-based anomaly detection methods, including MTAD-GAT \cite{zhao2020multivariate}, GDN \cite{deng2021graph}, GRELEN \cite{zhang2022grelen}, Anomaly Transformer \cite{xu2022anomaly}, NLinear \cite{zeng2023transformers}, DLinear \cite{zeng2023transformers}, DCdetector \cite{yang2023dcdetector}, PatchTST \cite{nie2023a}, DyEdgeGAT \cite{zhao2024dyedgegat}, iTransformer \cite{liu2024itransformer}, ModernTCN \cite{donghao2024moderntcn}, CATCH \cite{wu2025catch}, and GCAD \cite{liu2025gcad}. 
Details on the baselines are given in the Appendix.

\begin{table*}
\begin{centering}
\par\end{centering}
\centering{}\resizebox{1\textwidth}{!}{%
\begin{tabular}{lcccccccc}
\toprule 
 & \multicolumn{2}{c}{SWAT} & \multicolumn{2}{c}{WADI} & \multicolumn{2}{c}{PSM} & \multicolumn{2}{c}{TEP} \tabularnewline

 & AUC-ROC & AUC-PR & AUC-ROC & AUC-PR & AUC-ROC & AUC-PR & AUC-ROC & AUC-PR \tabularnewline
 \midrule 
 GSLAD  & 0.8607 & 0.7612 & 0.7018 & 0.5157 & 0.8193 & 0.7038 & 0.9563  & 0.9687\tabularnewline
 w/o structural deviation & 0.7449 & 0.2052 & 0.5925 & 0.2263 & 0.6585 & 0.3905 & 0.9169 & 0.9293 \tabularnewline
 w/o Phase II training & 0.7561 & 0.2170 & 0.6805 & 0.4781 & 0.7721 & 0.5344 & 0.9494 & 0.9620 \tabularnewline
 w/o uncertainty awareness & 0.8605 & 0.7630 & 0.6595 & 0.1217 & 0.7478 & 0.5874 & 0.9124 & 0.9199 \tabularnewline
 
w/o condition awareness & 0.8489 & 0.7099 & 0.6756 & 0.4511 & 0.7785 & 0.5909 & 0.9405  & 0.9509\tabularnewline
\bottomrule
\end{tabular}}
\caption{Ablation Study.}
\label{tab:ablation}
\end{table*}
\subsubsection{Implementation Details}
Each dataset consists of two parts: unlabeled data under normal working conditions and labeled data containing some anomalies.
We use 80\% of the unlabeled normal data for training, and the remaining 20\% is used for validation. Testing is conducted on the labeled data containing anomalies.
The data is standardized using the mean and standard deviation computed from the normal training split, and the resulting time series is segmented into sliding windows. 
Since most methods do not provide a way to set predetermined thresholds, we evaluate using two threshold-independent metrics: the area under the curve (AUC) of the Receiver Operating Characteristic (ROC) \cite{fawcett2006introduction} and the Precision-Recall Curve (PRC) \cite{davis2006relationship}. 
More implementation details are provided in Appendix.

\subsection{Multivariate Time Series Anomaly Detection}
The anomaly detection results of the proposed GSLAD model and thirteen baseline models on four industrial anomaly detection benchmarks are summarized in Table \ref{tab:mtsad_results}.
GSLAD achieves the best ROC-AUC on SWaT, PSM, and TEP, and the best PR-AUC on all four datasets.
These results indicate that GSLAD consistently achieves state-of-the-art performance in most cases.
iTransformer performs well on two water network datasets, obtaining second-best results on three metrics. However, it fails to obtain competitive performance on PSM and TEP, especially on TEP where it obtains the worst results among all the baselines. 
GSLAD does not achieve the best result on every metric. On WADI, GCAD obtains a higher ROC-AUC.
Nevertheless, GSLAD remains competitive on these metrics and provides the strongest overall performance across the four datasets.

\subsection{Ablation Study}
To have a better understanding on how each individual component in GSLAD contribute to the improved performance, we conduct ablation studies in this section.
Specifically, we consider four variants of GSLAD by removing individual components from the model: 1) 'w/o structural deviation', which uses only the predictive deviation for anomaly scoring; 2) 'w/o Phase II training', which trains the model only with the Phase I predictive objective $\mathcal{L}_\mathrm{pred}$ and without the Phase II prototype-regularized refinement; 3) 'w/o uncertainty awareness', which removes the use of $\boldsymbol{\sigma}_k$ as uncertainty weighting and computes the unweighted structural deviation defined as \begin{equation}
    d_k(\mathbf{S}_t)
    =
    \frac{1}{N^2}
    \sum_{i=1}^N \sum_{j=1}^N
    (\mathbf{s}_{t,ij} - \mu_{k,ij})^2,
\end{equation} during both Phase II refinement training and inference stage; 4) 'w/o condition awareness', which does not modulate the query projection and key projection with the global state vector.
Table \ref{tab:ablation} summarizes the results for these four variants of GSLAD.

From the results, we observe that all four individual components of GSLAD contribute positively to the anomaly detection performance. 
Among them, removing the structural deviation from anomaly scoring leads to the largest overall degradation, confirming that structural deviation provides the critical anomaly evidence in GSLAD.
The performance drop is particularly significant on SWaT, WADI, and PSM, indicating that predictive deviation alone is insufficient to distinguish anomalies in such cases.
Removing Phase II training also substantially degrades performance, especially on SWaT and PSM. 
This observation validates that the more compact and stable structural patterns facilitated by prototype-regularized refinement can help anomaly detection.
Replacing the uncertainty-aware distance with an unweighted one consistently reduces performance, with especially large decreases on WADI, PSM, and TEP. This result demonstrates that deviations on stable edges provide stronger anomaly signals than deviations on edges that naturally fluctuate during normal operating regimes.
\begin{figure}
    \centering
    \includegraphics[width=1.0\columnwidth]{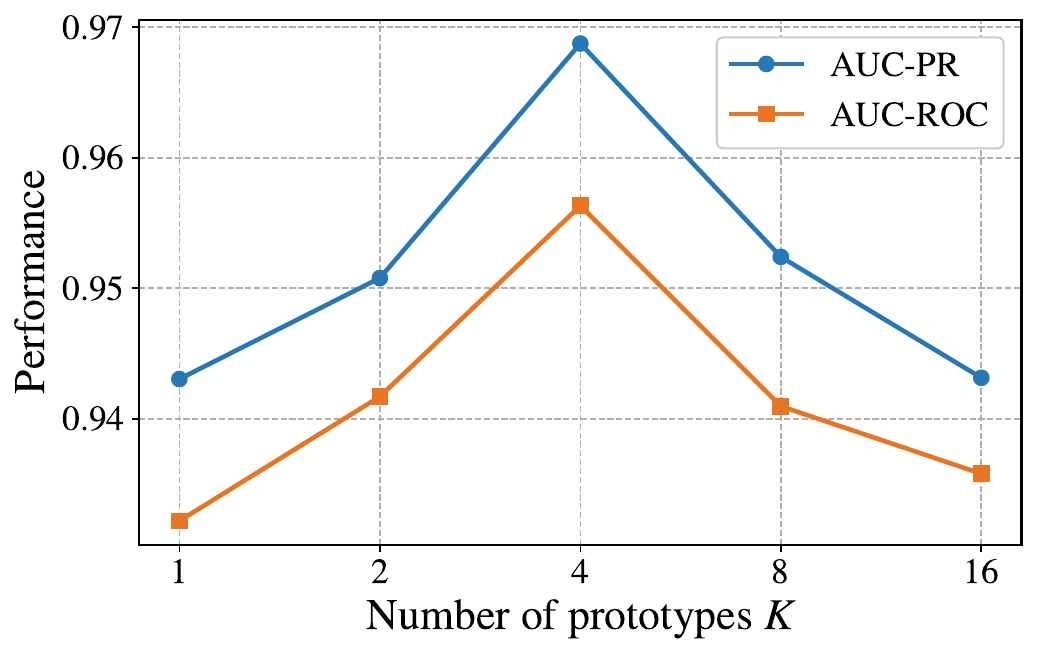}
    \caption{Sensitivity to prototype number $K$.}
    \label{fig:sensitivity}
\end{figure}
\subsection{Sensitivity analysis}
The number of prototypes used in GSLAD is an important hyperparameter for obtaining effective anomaly evidence.
We perform a sensitivity analysis of the model performance with respect to the prototype number $K$ using the TEP dataset.
The results are visualized in Figure \ref{fig:sensitivity}.
The best result is obtained when $K=4$. 
The results show that GSLAD performs consistently well when the number of prototypes is within a reasonable range close to the optimal prototype number.
Specifically, with $K=2$ or $K=8$, compared with the second-best baseline model in Table \ref{tab:mtsad_results}, GSLAD still obtains better AUC-ROC and similar AUC-PR.
However, with an overly large or small number of prototypes, the model performance becomes worse than other baselines.
This is because too many prototypes make the clusters and statistics inaccurate and unstable, while too few prototypes are not enough to cover all the different normal operating regimes and may cause prototypes that correspond to no actual operating regime.


\begin{figure}
    \centering
    \includegraphics[width=1.0\columnwidth]{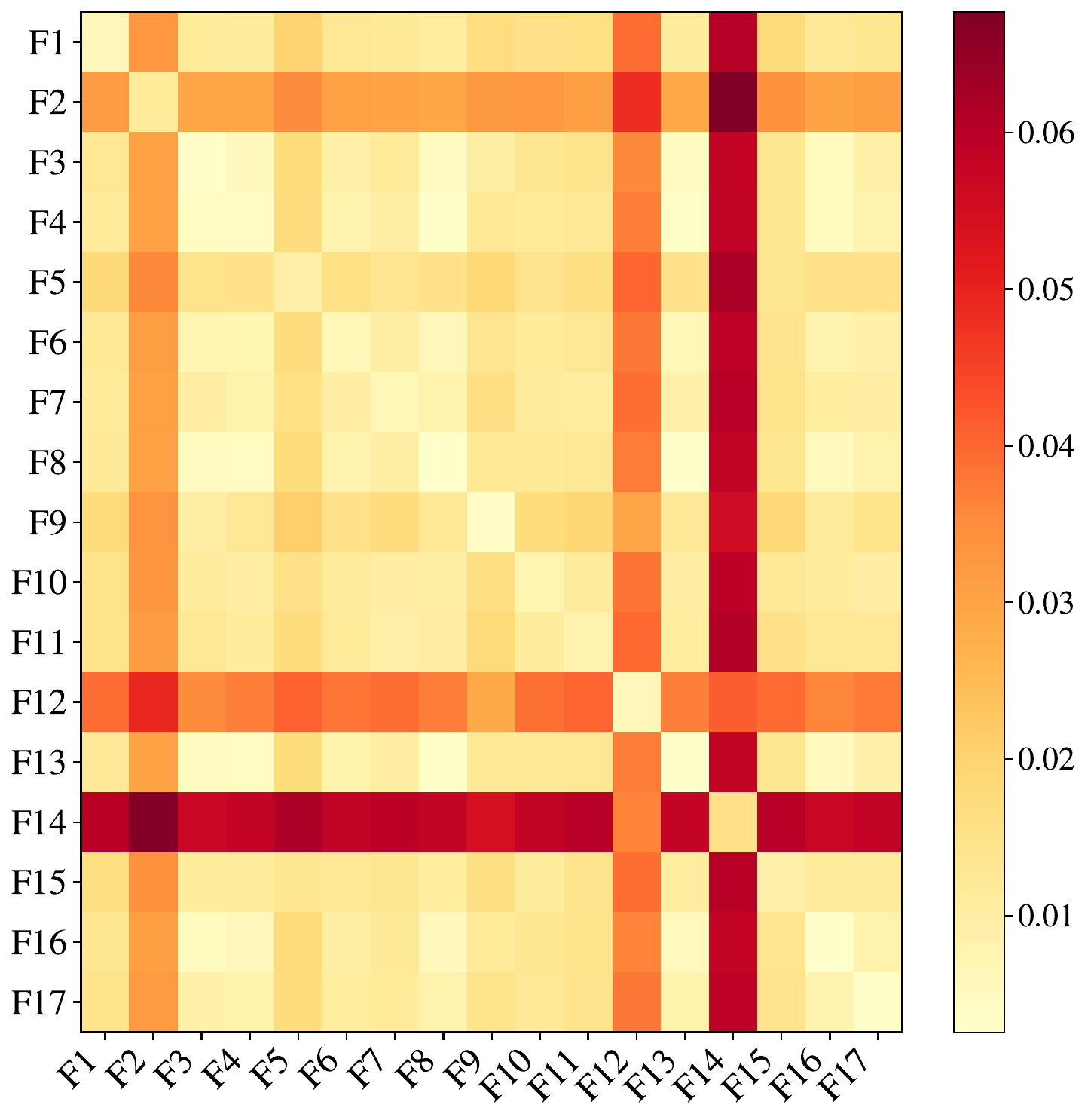}
    \caption{Cross-fault structural matrix.}
    \label{fig:cross_fault_heatmap}
\end{figure}
\subsection{Cross-Fault Structural Analysis}
In this section, we examine whether the learned structural space captures fault-specific structural patterns.
This analysis is conducted on TEP because it contains 17 different fault types covering different fault mechanisms with annotations.
Note that these labels are used only for post-hoc interpretation and are never used during model training or anomaly scoring.
For each fault type $i$, we first construct a class-level reference graph as follows:
\begin{equation}
    \mathbf{F}_i=\frac{1}{|\mathcal{M}_i|}\sum_{m\in \mathcal{M}_i}\mathbf{S}^{(m)},
\end{equation}
where $\mathcal{M}_i$ denotes the set of windows belonging to fault type $i$ and $\mathbf{S}^{(m)}$ is the inferred graph of window $m$.
We then compute the mean distance from samples of fault type $j$ to the reference graph of fault type $i$ as follows:
\begin{equation}
C_{ij}=\frac{1}{|\mathcal{M}_j|}\sum_{m\in \mathcal{M}_j}\|\mathbf{S}^{(m)}-\mathbf{F}_i\|.
\end{equation}
We visualize the cross-fault structural matrix in Figure \ref{fig:cross_fault_heatmap}.
The matrix exhibits a clear diagonal pattern that all the diagonal entries are the minimum in the corresponding rows and columns.
On average, the within-class distance is 34.1\% lower than the corresponding nearest competing class.
This observation indicates that the inferred graphs are generally closest to the reference graph of the same fault type.
Thus, we can conclude that empirically, the inferred graph generated with GSLAD would deviate in different directions for different fault types, despite that the model is trained without fault type supervision.
Among all the faults, F2, F12, and F14 exhibit particularly large margins from competing classes, suggesting that they induce most distinctive changes in structural patterns. Overall, the analysis shows that graph structures learned by GSLAD provide information beyond binary anomaly detection and may support fault characterization.

\subsection{Structural Deviation Interpretation}
To investigate whether structural deviation can support fault diagnosis such as root-cause localization, we analyze Fault 14 as an example, which represents a reactor cooling-water valve-sticking fault that mainly affects the reactor cooling-water subsystem.
For each anomalous window, we first compute the uncertainty-normalized edge-deviation matrix with respect to the closest normal prototype:
\begin{equation}
    D_{ij}
    =\min_{k \in \{1,\ldots,K\}}
    \frac{
    (s_{ij} - \mu_{k,ij})^2
    }{
    \sigma_{k,ij}^2 + \sigma_0^2
    }.
\end{equation}
We then define the node-level deviation score as the average of its incoming and outgoing edge deviations:
\begin{equation}
\rho_v=
\frac{1}{2}
\left(
\frac{1}{N}\sum_jD_{vj}
+
\frac{1}{N}\sum_iD_{iv}
\right).
\end{equation}
Figure 4 presents the variables with largest $\rho_v$.
The three most affected nodes are XMV(10), XMEAS(9), and XMEAS(21), corresponding to reactor cooling-water flow, reactor temperature, and cooling-water outlet temperature, respectively. These variables belong to the subsystem directly associated with Fault 14, and the third-ranked variable has a deviation more than 6.4 times larger than the fourth-ranked variable. This concentration of node-level deviation suggests that the uncertainty-aware structural deviation provides a useful signal for localizing the root cause.
\begin{figure}
    \centering
    \includegraphics[width=1\columnwidth]{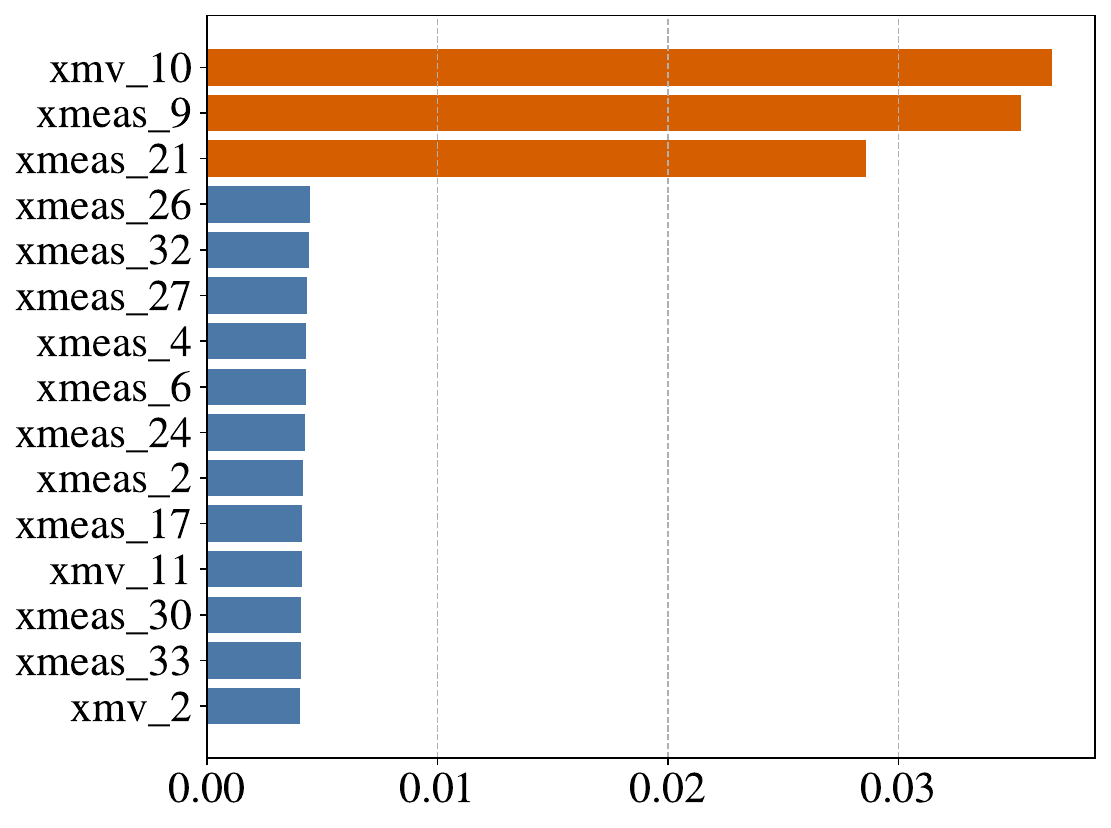}
    \caption{Node-level deviation ranking.}
    \label{fig:cross_fault_heatmap}
\end{figure}

More experiments on runtime analysis and anomaly score time series analysis are provided in the Appendix.
\section{Conclusion}
In this paper, we introduced GSLAD, a prototype-regularized graph structure learning framework for unsupervised MTS anomaly detection. GSLAD learns window-dependent graphs and represents normal structural behavior with a bank of prototypes, enabling structural deviation to complement predictive deviation for anomaly scoring. Experiments demonstrate strong overall performance of GSLAD, while structural analyses suggest potential for fault characterization and root-cause localization. Future work will investigate graph structure learning with stronger causal interpretability for industrial anomaly detection.
\newpage
\bibliography{aaai2027}

\newpage
\newpage
\appendix
\onecolumn
\section{Appendix}
\subsection{Baselines}
To evaluate the performance of our proposed GSLAD method, we compare it with various baselines. We briefly introduce these baselines below:
\begin{itemize}
    \item \textbf{MTAD-GAT} \cite{zhao2020multivariate} employs parallel feature- and temporal-oriented graph attention layers and jointly optimizes forecasting and reconstruction objectives for multivariate time-series anomaly detection.

    \item \textbf{GDN} \cite{deng2021graph} learns a sparse dependency graph among sensors through node embeddings and graph attention, and detects anomalies based on forecasting deviations from the learned normal behavior.

    \item \textbf{GRELEN} \cite{zhang2022grelen} integrates a variational autoencoder with stochastic graph relational learning to capture inter-sensor dependencies and construct a relation-aware anomaly score.

    \item \textbf{Anomaly Transformer} \cite{xu2022anomaly} introduces anomaly attention and a minimax learning strategy to distinguish anomalies through the discrepancy between prior and series associations.

    \item \textbf{NLinear} \cite{zeng2023transformers} mitigates distribution shifts by subtracting the last observed value before linear forecasting and adding it back to the prediction.

    \item \textbf{DLinear} \cite{zeng2023transformers} decomposes each time series into trend and seasonal components and forecasts them using separate linear mappings.

    \item \textbf{DCdetector} \cite{yang2023dcdetector} adopts multi-scale dual-attention contrastive learning to obtain discriminative representations without relying on a reconstruction objective.

    \item \textbf{PatchTST} \cite{nie2023a} divides each time series into temporal patches and processes different variables independently with a weight-shared Transformer encoder.

    \item \textbf{DyEdgeGAT} \cite{zhao2024dyedgegat} dynamically infers evolving inter-sensor edges and incorporates operating-condition context into graph-attention-based signal reconstruction.

    \item \textbf{iTransformer} \cite{liu2024itransformer} represents individual variables as tokens, using self-attention to capture inter-variable dependencies and feed-forward networks to learn temporal representations.

    \item \textbf{ModernTCN} \cite{donghao2024moderntcn} modernizes temporal convolutional networks with large-kernel depthwise convolutions and convolutional feed-forward blocks to capture long-range temporal patterns efficiently.

    \item \textbf{CATCH} \cite{wu2025catch} partitions signals into frequency-domain patches and employs masked channel fusion to model fine-grained spectral characteristics and channel correlations.

    \item \textbf{GCAD} \cite{liu2025gcad} dynamically extracts Granger-causality graphs from predictor gradients and detects anomalies through deviations in the learned causal patterns.
\end{itemize}
For forecasting models, anomaly scores are computed from forecasting deviations under the same evaluation protocol. Similar to GCAD, there is another recent work that infers the graph for anomaly detection based on intermediate embeddings \cite{cho2025structured}. We do not include it for comparison since there is no public code available. For methods implemented in \cite{wu2025catch}, we use their implementation. For the other methods we use their publicly available official implementations.

\subsection{Implementation Details}
All the experiments are conducted on a NVIDIA A100 80G GPU.
For all the experimental results, we give the average performance and standard deviation with 5 independent trials. 
For all the datasets, we select windows of length 128.
The Adam optimizer is used in all experiments for model training \cite{KingBa15}.
We fix the training epochs of Phase I to 30 and the training epochs of Phase II to 10.
We use a batch size of 256 for model training.
The models’ hyperparameters are tuned based on the results of the validation set. 
The search space of hyperparameters are as follows:  1) horizon: \{1,4,8\}; 2) GNN layers: \{2, 4\}; 3)
embedding dimension: \{64, 128, 256\}; 4) weight parameter $\lambda$ in the loss function: \{1, 3, 10, 30, 100\}; 5) number of prototypes: \{1, 2, 4, 8\}.
The learning rate and weight decay are set to 1e-4 and 1e-5, respectively.
For generating the sparse graphs, we keep the top-$k$ outgoing neighbors for each node with $k=5$.
The temperature parameter $\tau$ for soft nearest prototype graph assignment during Phase II training is set to 0.05.
The global stabilization term $\sigma_0$ used for computing uncertainty-aware graph deviation is set to 0.05.
During inference, we use the sum of $z$-score normalized structural deviation and predictive deviation as the anomaly score, where we use the mean and standard deviation computed from training set to perform $z$-score normalization.

\subsection{Runtime Comparison}
\begin{figure}
    \centering
    \includegraphics[width=1\columnwidth]{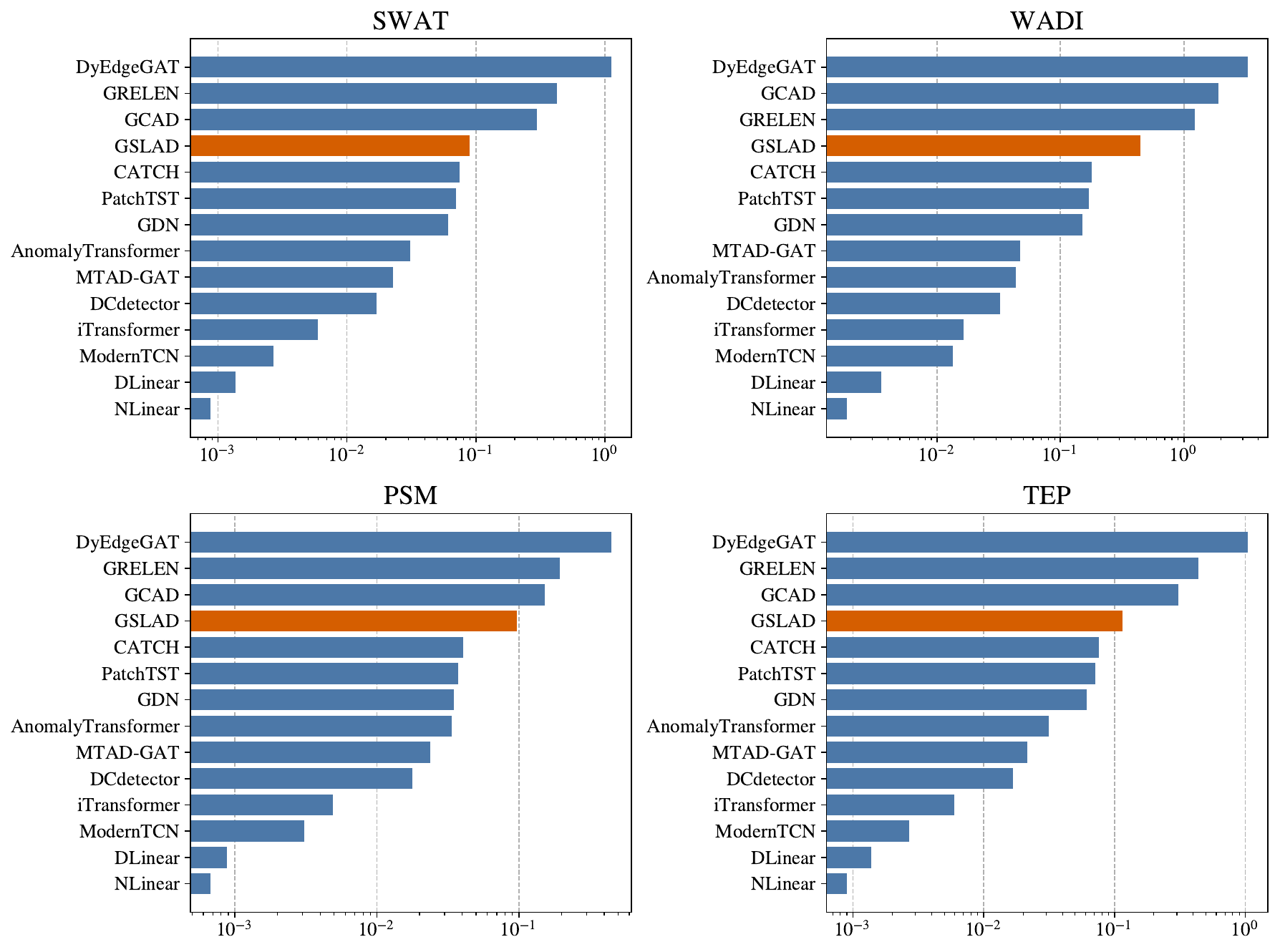}
    \caption{Per-sample inference time comparison.}
    \label{tab:runtime}
\end{figure}

Since GSLAD involves additional computation for graph structure learning, we perform a runtime comparison to evaluate the efficiency of GSLAD.
Specifically, we evaluate the per-sample inference time (in seconds) on four industrial datasets.
The results are shown in Figure \ref{tab:runtime}.
From the results, we can see that the models that jointly learn graph structures generally require longer inference time.
However, we observe that GSLAD remains substantially more efficient than the other baselines that also perform structure learning, including DyEdgeGAT, GCAD, and GRELEN.
For example, on the SWAT dataset, GSLAD is 12.5 times, 3.3 times, and 4.8 times faster than DyEdgeGAT, GCAD, and GRELEN, respectively.

\subsection{Anomaly Score Time Series Analysis}
\begin{figure}[t]
\centering
    \includegraphics[width=0.8\columnwidth]{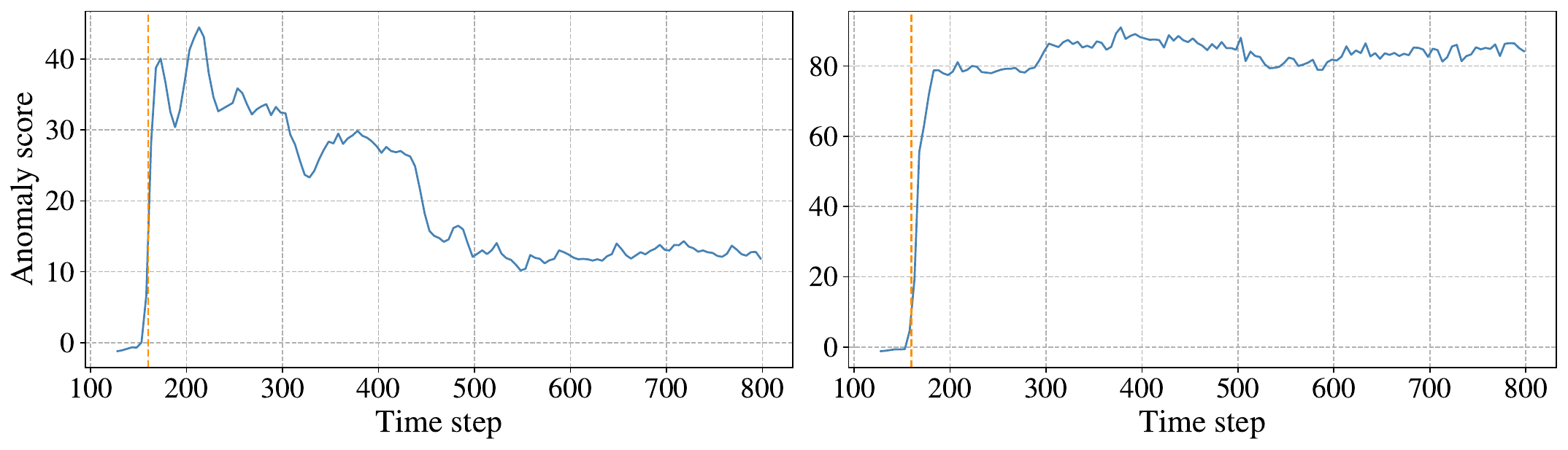}

    \caption{Anomaly score time series. The orange line represents the start time of the anomaly case.}
    \label{fig:anomaly_scores}
\end{figure}

In this section, We qualitatively examine how the anomaly score evolves around fault onset.
Figure \ref{fig:anomaly_scores} visualizes the anomaly scores time series for two faults in TEP, with the anomaly score containing both uncertainty-aware structural deviation and predictive deviation. Note that GSLAD treats each window independently and computes the anomaly score for each window separately. Both examples show a clear and sustained increase on anomaly score immediately after fault onset, indicating GSLAD provides strong anomaly signals. 
\end{document}